\documentclass{article}
\usepackage[utf8]{inputenc} 
\usepackage{geometry}
\usepackage{amsmath}
\usepackage{amssymb}
\usepackage{graphicx}
\usepackage{xcolor}
\usepackage{hyperref}
\usepackage{mathtools}
\usepackage{changepage} 
\usepackage{multirow}
\newcommand{\LinedItem}[1]{\item \begin{tabular}[t]{|p{17cm}}#1\end{tabular}}

\usepackage{palatino}

\DeclarePairedDelimiter\floor{\lfloor}{\rfloor}
\title{Investigating the Relationship Between Dropout \\Regularization and Model Complexity in Neural Networks}
\author{Jai Sharma, Milind Maiti, Christopher Sun}
\date{August 2021} 

\begin{document}
\maketitle

\begin{center}
\section*{Abstract}
\end{center}
Dropout Regularization, serving to reduce variance, is nearly ubiquitous in Deep Learning models. We explore the relationship between the dropout rate and model complexity by training 2,000 neural networks configured with random combinations of the dropout rate and the number of hidden units in each dense layer, on each of the three data sets we selected. The generated figures, with binary cross entropy loss and binary accuracy on the z-axis, question the common assumption that adding depth to a dense layer while increasing the dropout rate will certainly enhance performance. We also discover a complex correlation between the two hyperparameters that we proceed to quantify by building additional machine learning and Deep Learning models which predict the optimal dropout rate given some hidden units in each dense layer. Linear regression and polynomial logistic regression require the use of arbitrary thresholds to select the cost data points included in the regression and to assign the cost data points a binary classification, respectively. These machine learning models have mediocre performance because their naive nature prevented the modeling of complex decision boundaries. Turning to Deep Learning models, we build neural networks that predict the optimal dropout rate given the number of hidden units in each dense layer, the desired cost, and the desired accuracy of the model. Though, this attempt encounters a mathematical error that can be attributed to the failure of the vertical line test. The ultimate Deep Learning model is a neural network whose decision boundary represents the 2,000 previously generated data points. This final model leads us to devise a promising method for tuning hyperparameters to minimize computational expense yet maximize performance. The strategy can be applied to any model hyperparameters, with the prospect of more efficient tuning in industrial models. 

\clearpage

\section*{Introduction} 

Dropout Regularization is a technique used in Deep Learning and other branches of Machine Learning to combat overfitting on training data with respect to the validation set \cite{dropout}. Another hyperparameter, the architecture of a neural network, determines the complexity of the functions the model will be able to learn, with a deeper model that contains more hidden units able to reduce bias in the network \cite{DNN}. 
\\\\
Superficially, the implementation of dropout regularization seemingly counteracts the effect of building a model with more hidden units. While increased model complexity mitigates bias, higher dropout rates serve to restore bias to prevent overfitting. Thus, it is crucial for the Machine Learning practitioner to equilibrate these two hyperparameters to find the tuning that induces the finest performance. The common approach is to build a complex model accompanied by substantial regularization. However, the problem with this attempted one-size-fits-all approach is that training the model can require extensive computational resources and prolonged train times, not to mention the greater issue of suboptimal convergence.
\\\\
Thus, we seek to identify the relationship between the dropout rate and the number of hidden units in a neural network. We aim to demonstrate brief hyperparameter tuning with respect to cost and accuracy, specifically intending to illustrate that adding complexity in a neural network is not always the steadfast solution to unlock a better-performing model. Rather, a simpler model, which can be trained in much less time, can offer similar performance. We experiment with the relationship between variance and bias by studying model complexity and dropout regularization, showing that multiple model configurations can result in similar performance. In addition, we explore a method to find the best dropout rate to use in a Deep Learning model given the model complexity of each dense hidden layer. Finally, we share a generalized version of this insightful procedure that Machine Learning practitioners can utilize to more quickly perform hyperparameter tuning to balance bias and variance while optimizing validation cost and accuracy. 

\subsection*{Data Availability Statement}
For experimentation, we used three publicly available cardiovascular disease data sets. Each of the training examples in these three data sets represents individual patients. Medical information about the patients and a binary diagnosis label for cardiovascular disease are included for each data example. We selected these data sets because they are all relatively small, allowing for faster hyperparameter tuning. The following describes each of the data sets used (\textit{data sets will henceforth be referred to by the names below in bold}):

\paragraph{High-Level Data Set} The High-Level Data Set is available on the Kaggle \cite{kaggle_dataset} database and contains 70,000 examples and 11 high-level features, most binary and ternary. Prior to experimentation, this data set underwent slight preprocessing to remove statistical outliers using the 1.5 x IQR Rule with a modified coefficient of 2.5, which we believed better represented the spread of the data set. Training examples with features beyond the range permitted by the IQR Rule were excluded from the data set.  

\paragraph{Cleveland Data Set} The Cleveland Data Set is a famous source of cardiovascular disease data containing 303 examples and 75 features available on the UCI Machine Learning Repository \cite{uci}. Since only 14 of the 75 features are used by Machine Learning researchers when studying cardiovascular disease, we also experimented with this subset of features.

\paragraph{Combined Data Set} We assembled a data set that is a fusion of cardiovascular disease data sourced from hospitals in Cleveland, Hungary, Switzerland, and Long Beach, California. The individual data sets are available on the UCI Machine Learning Repository \cite{uci}. The Combined Data Set contains 1,025 examples with the same 14 features included in the Cleveland Data Set. 


\section*{Methods}
\subsection*{Measuring Performance of Deep Learning Models Configured With Unique Combinations of Dropout Rate and Layer Size}
To investigate the relationship between dropout regularization and model complexity, we explored hyperparameter tuning. We randomly chose combinations of model complexity and regularization, then trained 2,000 different neural networks \footnote{All Deep Learning models constructed during our research were trained using the highly optimized TensorFlow Keras library \cite{keras}.} using these randomly chosen hyperparameters to predict the cardiovascular disease diagnosis label given the medical information of the patients. Finally, we determined the binary cross entropy cost and binary accuracy of each model after 150 iterations of training. The architecture of the networks comprised six dense hidden layers each with the same number of hidden units and the same dropout rate. However, each model trained used a unique number of hidden units and a unique dropout rate. 
\begin{adjustwidth}{0cm}{}
    \LinedItem{Let \textit{\textbf{x}} be the number of hidden units in each of the six hidden layers of a constructed Deep Learning model.}
    \LinedItem{Let \textit{\textbf{y}} be the dropout rate for this Deep Learning model.}\\
\end{adjustwidth}
The value of \textit{\textbf{x}} ranged from $2^3=8$ to $2^{10}=1024$. To sample as many models with $2^3=8$ to $2^4=16$ hidden units as models with $2^9=512$ to $2^{10}=1024$ hidden units, we chose \textit{\textbf{x}} based on a logarithmic scale. Specifically, $\textit{\textbf{x}}=\floor{2^c}$, where $c$ was a uniformly random number chosen from the interval $(3,10)$. (The number \textit{\textbf{x}} must be an integer, so the floor function was performed.) The value of \textit{\textbf{y}} was a uniformly random number chosen from the interval $(0,1)$.
\\
\\
\noindent All other hyperparameters remained constant during experimentation: each model's weights were initialized with Xavier initialization using the same random seed, and each model was optimized with Adam Optimization \cite{optimization} using a mini-batch \cite{minibatch} size of 128 examples, undergoing 150 epochs of training. 
\[\includegraphics[scale=0.35]{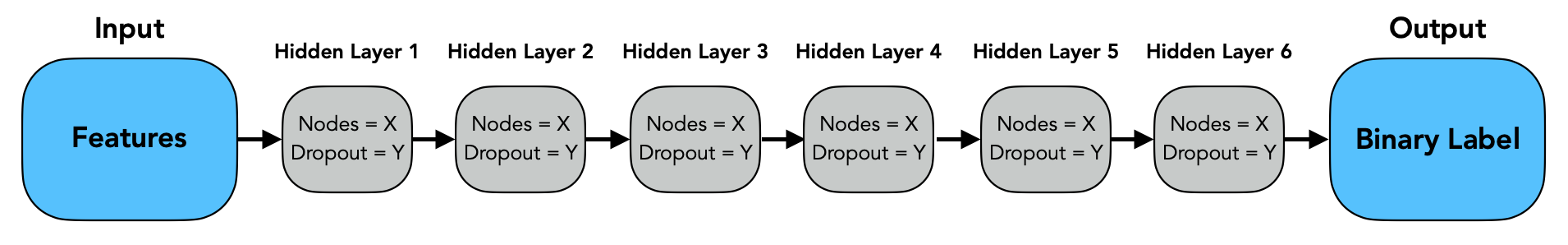}\]
\begin{align*}
    &\textbf{Figure 1: }\text{Generalized structure of each of the 2,000 neural networks built, where \textbf{X} is the number of }\\
    &\text{hidden units in each hidden layer and \textbf{Y} is the dropout rate.}
\end{align*}
After training 2,000 such models on each of the three data sets and recording the performance metrics of each model, we proceeded to plot all generated points on three-dimensional graphs: the number of hidden units on the $x$ axis, the dropout rate on the $y$ axis, and the attained cost/accuracy on the $z$ axis.

\[\hspace{-5mm}\includegraphics[scale=0.35]{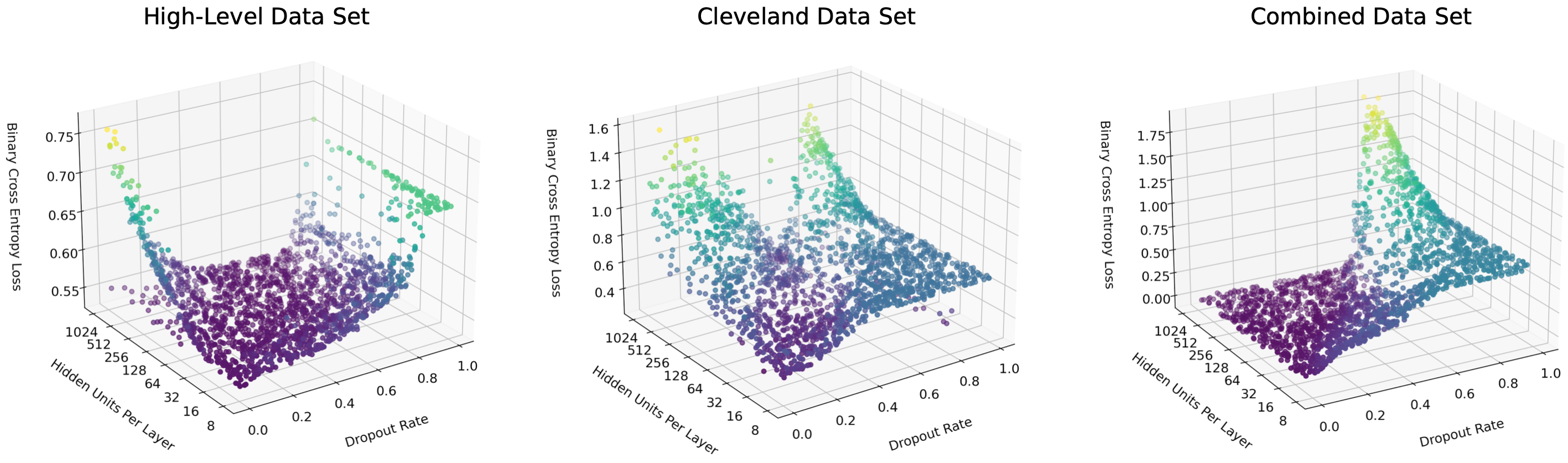}\]
\begin{align*}
    &\textbf{Figure 2: }\text{Cost plots generated by training 2,000 models on each of the three data sets, where the color map is }\\
    &\text{determined by the cost, the z-axis parameter. Notice that all plots have distinct contours that are able to be modeled.}
\end{align*}
\[\includegraphics[scale=0.34]{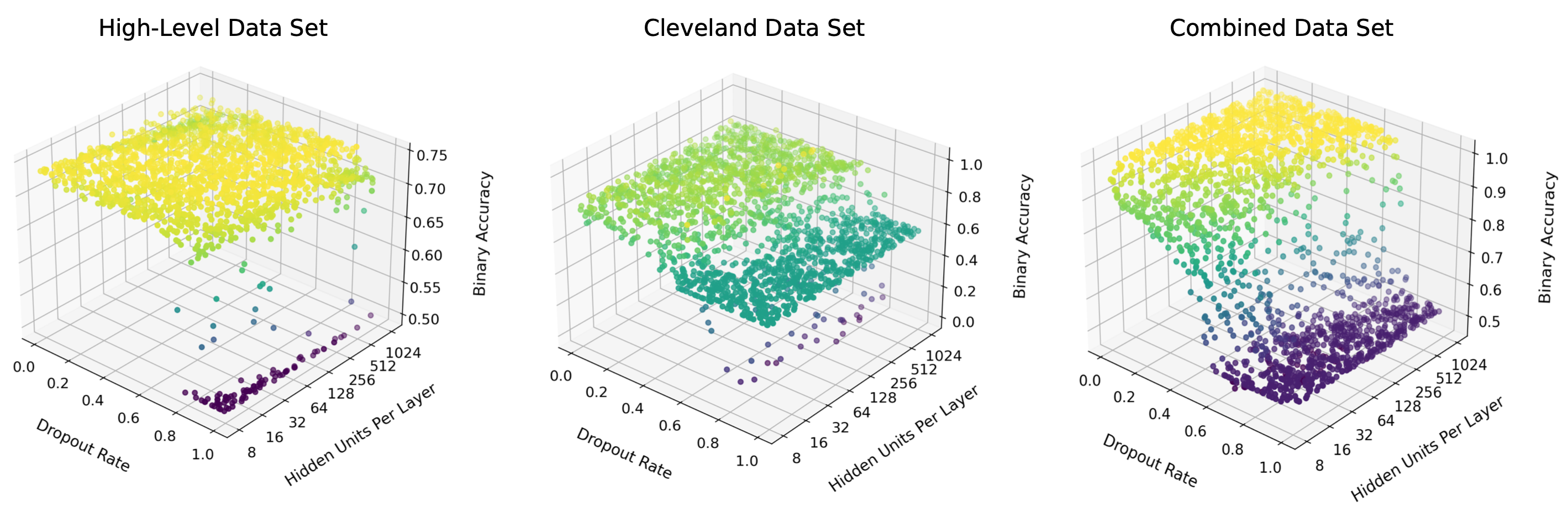}\]
\begin{align*}
    &\textbf{Figure 3: }\text{Accuracy plots generated by training 2,000 models on each of the three data sets, where the color map is }\\
    &\text{determined by the binary accuracy, the z-axis parameter. }
\end{align*}
\subsection*{Modeling a Generalized Relation Between Dropout Rate and Layer Size} 

Analyzing the figures above, we identified respective regions of lower cost and higher accuracy. For example, we observed a trough spanning the length of the cost plot of the High-Level data set that looked linear in nature. Considered together, our observations compelled us to answer the question: \textit{What is the correlation between dropout rate and model complexity that unlocks superior performance?} We hypothesized the optimal dropout rate to be a function of the number of hidden units in a hidden layer (assuming the number of hidden layers upon other variables stayed constant), and thus built a series of predictive models to fit the collected 2,000 points of data and quantify the hypothesized relationship. 

\paragraph{Linear Regression}
Linear Regression was the first technique utilized to fit points in the ``trough'' regions of all graphs. Initially, the boundaries of these troughs needed to be specified, requiring the manual setting of numerical and percentile thresholds of the 2,000 cost data points. Combinations of dropout rate and model complexity that achieved a cost lower than this numerical threshold or within this percentile threshold were included in the regression, while other inferior combinations were excluded. We experimented with numerical thresholds incrementing in 0.001 from 0.54 to 0.55 and percentile thresholds incrementing in 5\% from 5\% to 25\%. Clearly, the selection of thresholds was inherently subjective, not to mention that abundant amounts of usable data were being wasted because of the procedure's selective nature. 

\paragraph{Polynomial Logistic Regression}
To utilize all data points collected, we turned to Logistic Regression to assign models a binary classification based on their cost. Because cost is on a continuous scale, we transformed cost into a binary feature by labeling costs under a hand-chosen percentile threshold as 1 (superior) and all other costs as 0 (inferior). We tested 10\% and 25\% as these percentile thresholds, after which we built Logistic Regression models to classify combinations of dropout rate and number of hidden units depending on their achieved cost. Feeding second-order and third-order polynomial features into the models allowed for the learning of nonlinear decision boundaries. 
\\\\
Though the Logistic Regression models employed all 2,000 generated data points, the arbitrary selection of a percentile for the classification of costs was still required. In addition, the selection of the degree of polynomial features was not methodical. Furthermore, the Logistic Regression models were quite naive in the sense that they assigned the trained Deep Learning models just two elementary labels, which meant they were incapable of differentiating between a very poor model and a below average model or between sufficient model and an exceptional model.

\subsubsection*{Neural Networks}

\paragraph{1. Neural Networks for Optimal Dropout Rate Prediction} Deficiencies of previous machine learning models led us to utilize the complex hypothesis functions of neural networks \cite{DNN}. With these models, we wished to rigorously quantify the performance of model using the exact cost rather than using discrete labels. Accordingly, we constructed three neural networks \footnote{These neural networks are not to be confused with the networks trained on the cardiovascular disease data or the network that outputs a continuous dropout rate.} that received the number of hidden units, the desired cost, and the desired accuracy as inputs. At test time, the desired cost was interpreted as the minimum cost attained in our generated data set, and the desired accuracy was interpreted as the maximum accuracy attained in our generated data set. Thus, we were able to feed an array of hidden units ranging from $2^{3}$ to $2^{10}$ into these neural networks to obtain the optimal dropout rate for each value. Instead of yielding a binary label, the network yielded a continuous output between 0 and 1, representing the optimal dropout rate given the inputs. 
\[\includegraphics[scale=0.25]{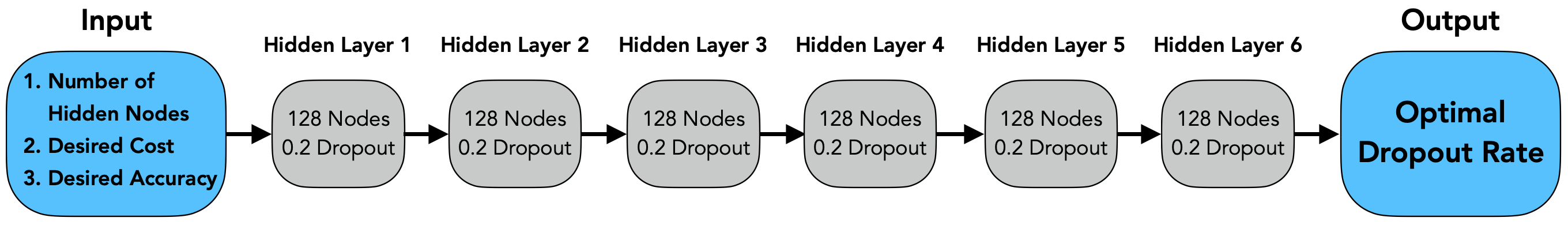}\]
\begin{align*}
    &\textbf{Figure 4: }\text{Generalized architecture of the neural network trained on the 2,000 generated points to predict optimal }\\
    &\text{dropout rate given a number of hidden units in each dense layer, the desired cost, and the desired accuracy. }
\end{align*}
\noindent The suboptimal performance of these neural networks gave us insight on how to implement the same type of model more effectively.

\paragraph{2. Neural Networks for Surface Plots}
\noindent We turned to a more rational approach, training neural networks to predict the cost of a Deep Learning model given the number of hidden units in the hidden layers and the dropout rate of the Deep Learning model. 
\\\\
Two neural networks were trained for each of the three data sets to model cost and accuracy, these networks containing six hidden layers with 16 hidden units each and regularized with a dropout rate of 0.1. We created three-dimensional surface plots which were representative of the continuous cost predictions given hyperparameter combinations. These decision boundaries were juxtaposed with the generated data points to visualize the models' performance. (These plots are shown in the Results section.)

\section*{Results}

\subsection*{Performance of Deep Learning Models With Unique Hyperparameter Combinations}

Analyzing the generated data led to various findings about the nature of the interactions between dropout rate, number of hidden units in a hidden layer, and the cost and accuracy achieved. Some figures, such as the accuracy plot of the Cleveland Data Set, depicted succinct regions in which the cost or accuracy was determined entirely by only a single hyperparameter of the two we analyzed. Other figures, such as the cost plot of the High-Level Data Set, showed that using a more hidden units with a higher dropout rate yielded the same performance as using less hidden units with a lower dropout rate. In contrast, still other figures, such as the accuracy plot of the Combined Data Set, favored the use of a lower dropout rate without regard for the number of hidden units.
\\\\
Such findings compelled us to rethink common presupposed relationships between the these two hyperparameters, challenging the frequent assumption that a larger number of hidden units in a dense layer simply requires a larger dropout rate to enhance performance. In addition, the plots discredited the notion that increasing the number of hidden units results in better performance. In fact, many spatial regions of the highest cost were achieved by models that had a relatively large amount of hidden units. We urge network designers to consider that an increase in model complexity, though theoretically reducing bias, can actually lower the performance of a model. The involuted balance created by regularization parameters such as the dropout rate cannot be readily oversimplified into a questionable cliche that adding hidden nodes unlocks superior performance. 
\subsection*{Generalized Relation Between Dropout Rate and Layer Size}
\paragraph{Linear Regression}
When examining the performance of the linear models which were fitted on a subset of the generated data to predict the optimal dropout rate, we noticed that while effective for the High-Level data set, linear models were unable to perform well on the Combined data set and the Cleveland data set, particularly due to the non-linearity of the problem and the disarray of the selected points. In addition, with the threshold percentile being set at the relatively low level of 25\%, 75\% of the data or 1500 points in our case were unused, amounting to waste in computational resources. Arbitrary selection of the threshold percentile was another undesirable byproduct of this model, as it was unclear which threshold percentile would lead to better performance.
\renewcommand{\arraystretch}{1.5}
\begin{center}
\begin{tabular}{|p{2.8cm}||p{3.3cm}|p{3.3cm}|p{3.3cm}|}
    \hline
    \multicolumn{4}{|c|}{\textbf{Mean Absolute Error of Linear Regression}}\\
    \hline
    & High-Level Data Set &Cleveland Data Set&Combined Data Set\\
    \hline
    10\% Threshold & 0.0609 & 0.0976 & 0.0768\\
    \hline
    25\% Threshold & 0.0942 & 0.1084 & 0.1137\\
    \hline
\end{tabular}
\end{center}
\[\hspace{-5mm}\includegraphics[scale=0.5]{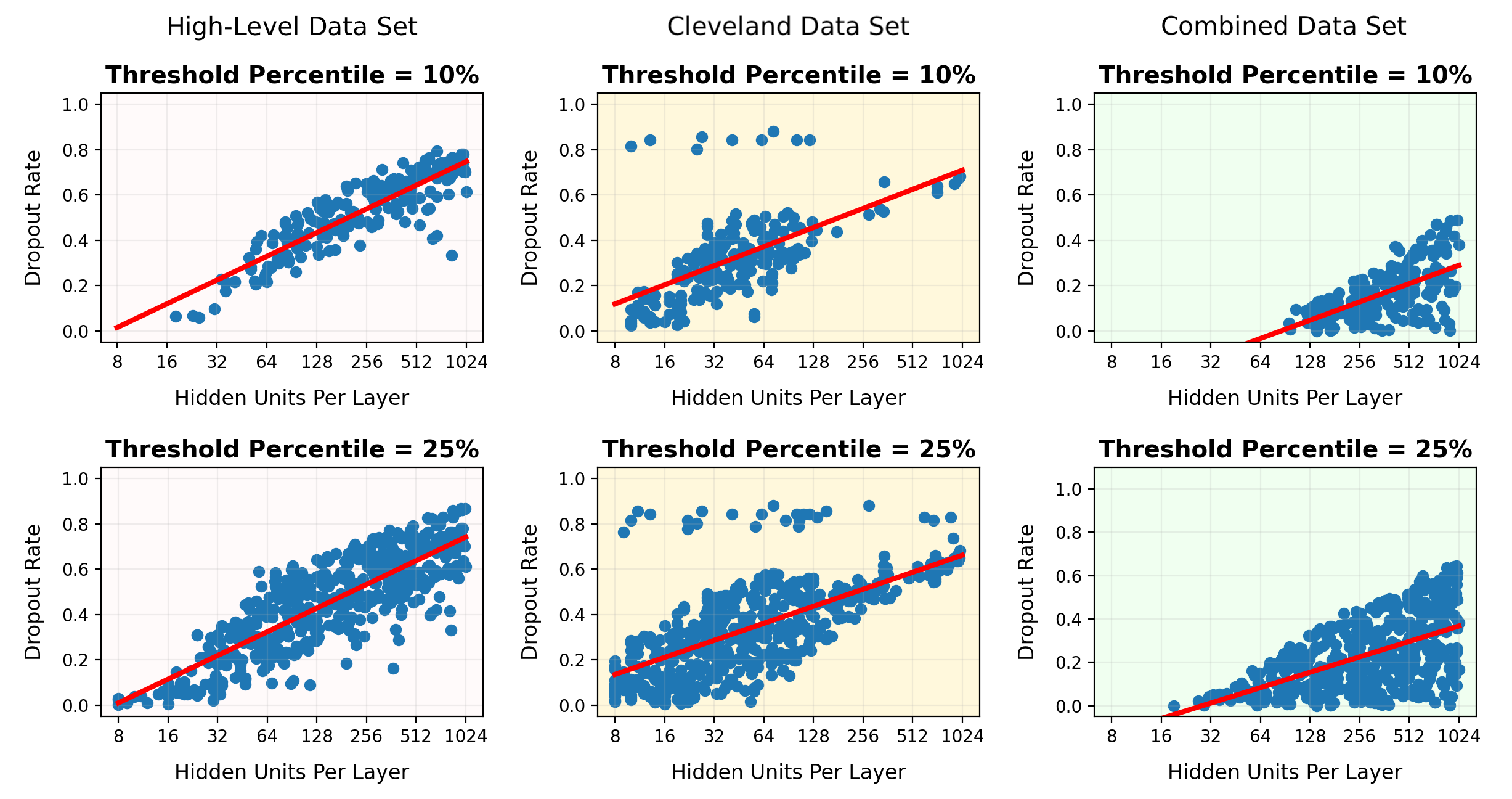}\]
\begin{align*}
    &\textbf{Figure 5: }\text{Visualizations of Linear Regression on all three data sets, with different percentile thresholds }\\
    &\text{being used to select points included in the regression. Notice that this approach had different levels of }\\
    &\text{effectiveness across data sets and performed especially poorly on the Combined Data Set.} 
\end{align*}
\paragraph{Polynomial Logistic Regression} The Logistic Regression models greatly improved upon the performance of the Linear Regression due to the extra data employed. When using Logistic Regression, the non-linearity of separating optimal models from non-optimal models became apparent, and thus we employed higher-order features. When utilizing these higher-order features, we found that including third-order features greatly improved performance without overfitting the data. Logistic Regression with higher order features yielded promising results for distinguishing ``superior'' models from ``inferior'' models. Nonetheless, Logistic Regression still fell short in that it offered a binary classification of the data. Due to the absence of clear definitions of optimal and non-optimal models, arbitrary thresholds had to be selected. The more desirable solution would be to utilize the continuous cost values. 
\renewcommand{\arraystretch}{1.5}
\begin{center}
\begin{tabular}{|p{3.9cm}||p{3.3cm}|p{3.3cm}|p{3.3cm}|}
    \hline
    \multicolumn{4}{|c|}{\textbf{Binary Accuracy of Polynomial Logistic Regression (25\% Threshold)}}\\
    \hline
    & High-Level Data Set &Cleveland Data Set&Combined Data Set\\
    \hline
    Second Degree Features & 0.8490 & 0.8695 & 0.9825\\
    \hline
    Third Degree Features & 0.9005 & 0.8870 & 0.9820\\
    \hline
\end{tabular}
\end{center}
\[\hspace{-5mm}\includegraphics[scale=0.5]{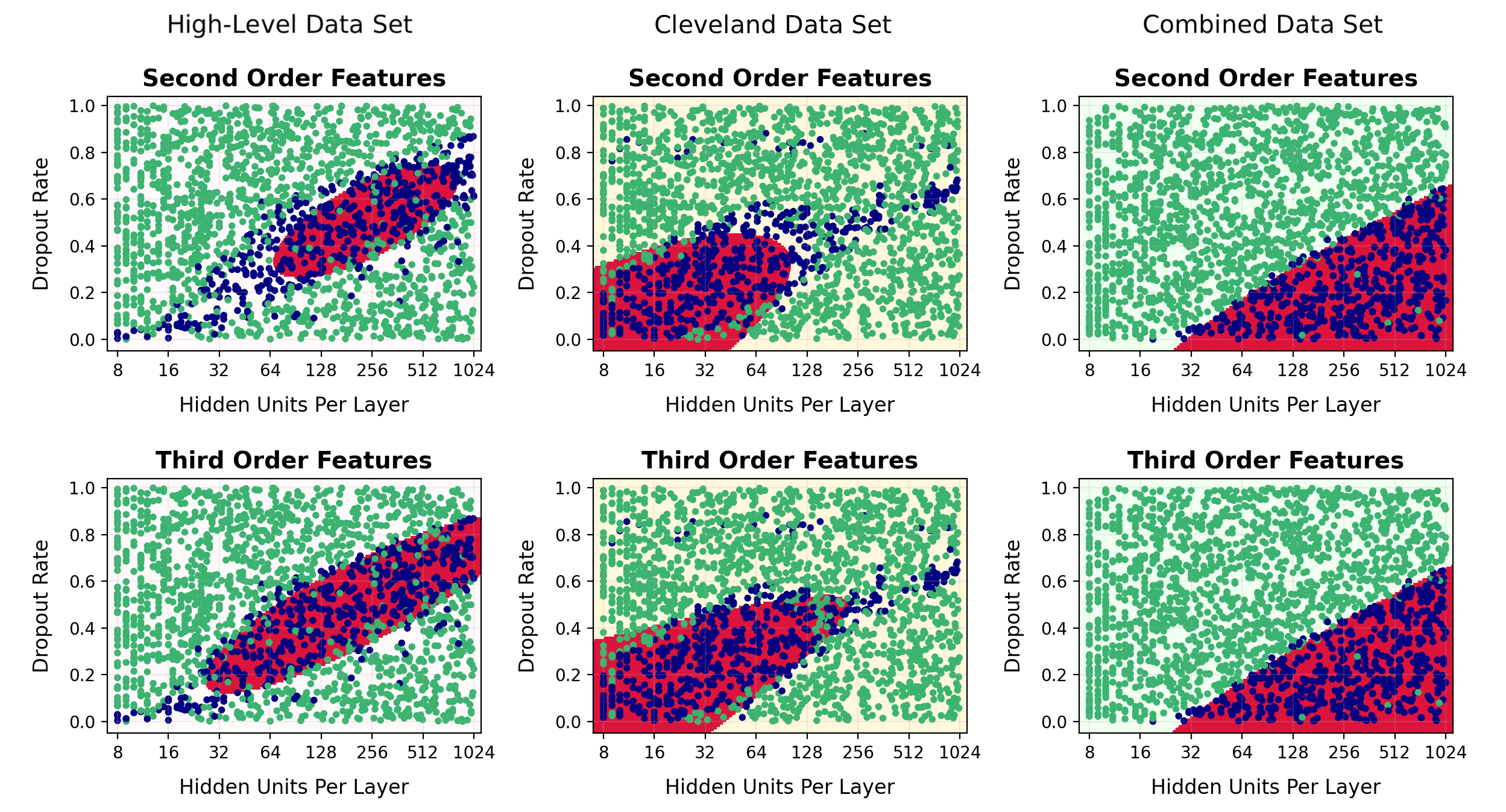}\]
\begin{align*}
    &\textbf{Figure 6: }\text{Visualizations of the Polynomial Logistic Regression decision boundaries on all three data sets, using a }\\
    &\text{percentile threshold of 25\%. Red areas are regions that the models classify as having a superior cost (``good'', 1), }\\
    &\text{while the background color represents the models' classification of inferior cost (``bad'', 0). Blue points symbolize } \\
    &\text{combinations of hyperparameters that achieve costs within the 25\% threshold, while green points }\\
     &\text{achieve costs that exceed the threshold.}
\end{align*}
\paragraph{Neural Networks for Optimal Dropout Rate Prediction} The relationship between the number of hidden units in each dense layer and the optimal dropout rate, as modeled by the first type of neural network introduced in the Methods section, is plotted below.
\renewcommand{\arraystretch}{1.5}
\begin{center}
\begin{tabular}{|p{3.4cm}|p{3.4cm}|p{3.4cm}|}
    \hline
    \multicolumn{3}{|c|}{\textbf{Mean Absolute Error of Neural Network Used to Predict Dropout Rate}}\\
    \hline
    High-Level Data Set &Cleveland Data Set&Combined Data Set\\
    \hline
    0.1384 & 0.0995 & 0.1640\\
    \hline
\end{tabular}
\end{center}
\[\includegraphics[scale=0.5]{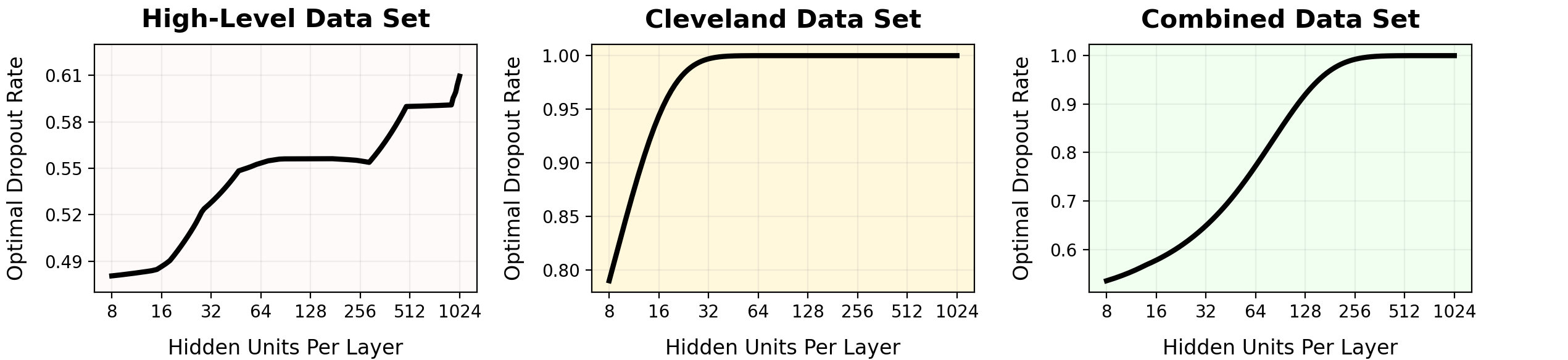}\]
\begin{align*}
    &\textbf{Figure 7: }\text{These figures plot the number of hidden units in each dense layer on the $x$ axis against the predicted }\\
    &\text{optimal dropout rate on the $y$ axis.}
\end{align*}
\noindent When we plotted the decision boundaries of the the neural networks in three dimensions, we saw that the models had disreputable performances, which caused us to speculate that this was due to the fundamental nature of the idea itself. We realized that theoretically, more than one dropout rate can be configured with the same number of hidden units to achieve the same cost and accuracy. Thus, we ventured to discover why the dropout rate is not always a function of the number of hidden units, cost, and accuracy. To explain this mathematically: 
\begin{adjustwidth}{0cm}{}
    \LinedItem{Let $n$ be the number of hidden units in each hidden layer for some Deep Learning model.}
    \LinedItem{Let $p$ be the optimal dropout rate given $n$.}
    \LinedItem{Let $\epsilon_1$ and $\epsilon_2$ both be some positive numbers.}\\
\end{adjustwidth}
By definition, the configuration of $p$ dropout rate achieves the lowest possible cost for all possible models with $n$ hidden units in each hidden layer. (Side Note: This configuration \textbf{is not} guaranteed to achieve the highest possible accuracy, only the lowest possible cost.)
\\\\
Using $p+\epsilon_1$ or $p-\epsilon_2$ as the dropout rate will result in a higher cost than using $p$ as the dropout rate. The cost function graph, though not appearing as completely continuous, strongly suggests that there exists some values for $\epsilon_1$ and $\epsilon_2$ such that using $p+\epsilon_1$ or $p+\epsilon_2$ as the dropout rate will result in the \textit{same} cost. (If not exactly the same, then our data indicates that $\epsilon_1$ and $\epsilon_2$ can be found such that the resulting costs have a negligible difference.) This observation explains why it is impossible to perfectly predict the optimal dropout rate given the number of hidden units in each of the hidden layers, the cost of the model, and the accuracy of the model.

\paragraph{Network Networks for Surface Plots} These neural networks were significant because they enabled us to test thousands of configurations of hyperparameters through forward propagation in a matter of seconds, demonstrating an immense reduction in time of the hyperparameter tuning process.\\
\renewcommand{\arraystretch}{1.5}
\begin{center}
\begin{tabular}{|p{4.1cm}||p{3.3cm}|p{3.3cm}|p{3.3cm}|}
    \hline
    \multicolumn{4}{|c|}{\textbf{Mean Absolute Error of Neural Networks for Surface Plots}}\\
    \hline
    & High-Level Data Set &Cleveland Data Set&Combined Data Set\\
    \hline
    Cost Points   & 0.0115    &0.0897&   0.1156\\
    \hline
    Accuracy Points&   0.0100  & 0.0459   &0.0479\\
    \hline
\end{tabular}
\end{center}

\[\includegraphics[scale=0.4]{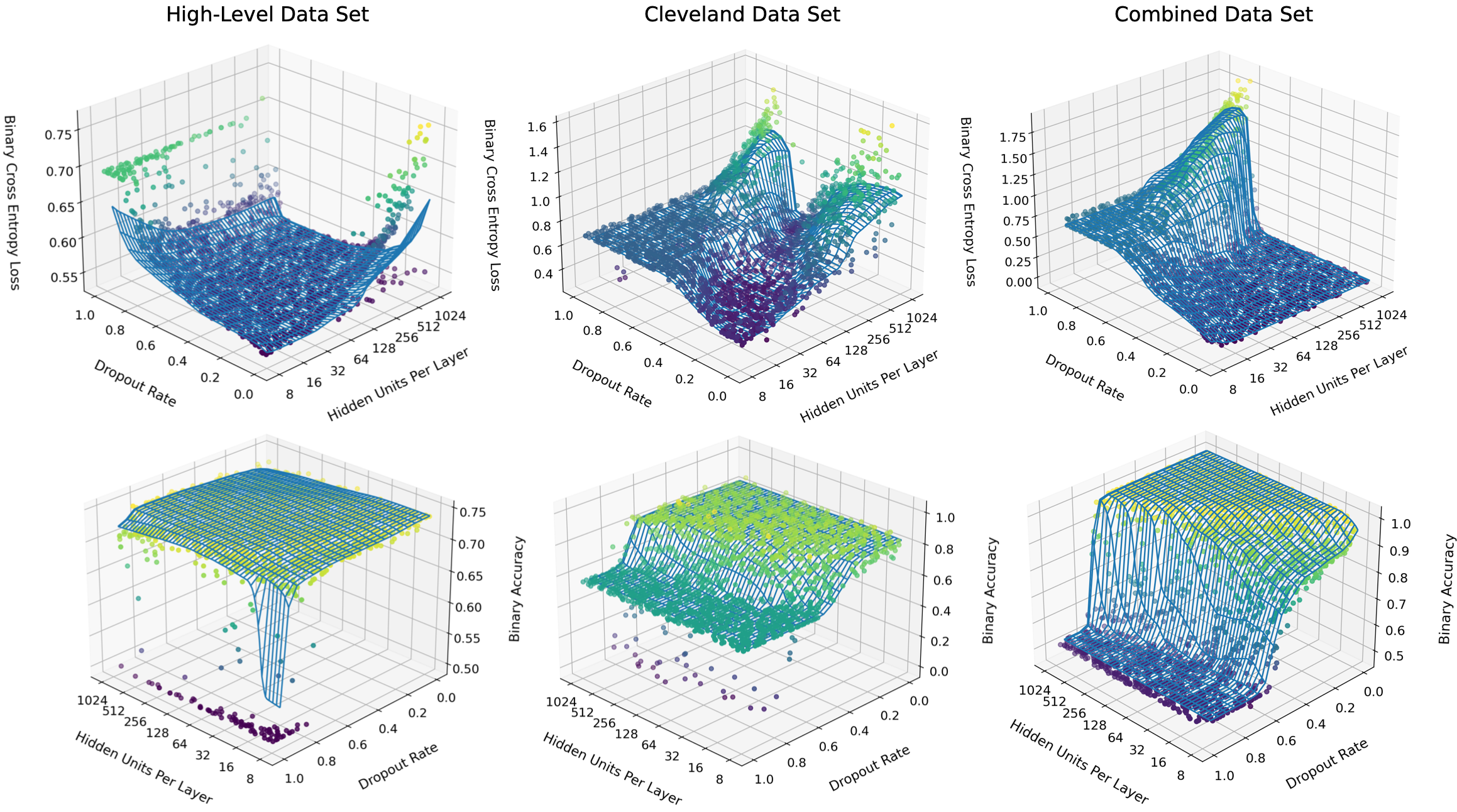}\]
\begin{align*}
    &\textbf{Figure 8: }\text{Visualization of the decision boundaries of the neural networks used to model the 2,000 generated }\\
    &\text{points, creating surface plots.}
\end{align*}

\section*{Discussion}
Allowing more flexibility in designing network architectures opens up many new avenues for the network designer. Our findings regarding the effects of regularization and hyperparameter tuning should motivate network designers to prioritize the conservation of valuable computational resources by designing more efficient architectures. 
\\
\\
Rapid prototyping in industrial applications can also be made more accessible when designers are not as inclined to tediously tune bigger models, but instead implement a larger volume of smaller models. While each neural network used in our research was small and was trained in roughly 10 minutes on GPU runtime, most industrial Deep Learning models utilize massive data sets with numerous features to tackle difficult problems (e.g. image recognition, sentiment analysis, self-supervised learning tasks). In these cases, training many models in a short period of time and iterating through the development process is essential to creating deployable products. Hence, the results of our research call for a more insightful approach to tune hyperparameters in neural networks. 

\subsection*{A Rapid and Effective Hyperparameter Tuning Method}
The following is an outline of a new method to accelerate the meticulous process of hyperparameter adjustment in Deep Learning models. 
\begin{adjustwidth}{0.4cm}{}
    \LinedItem{Let \textit{\textbf{n}} be the initial number of models trained on random hyperparameter settings chosen according to their respective range of values (in our case \textit{\textbf{n}} = 2,000; the number of hidden units in each dense layer was chosen on a logarithmic scale from $2^3$ to $2^{10}$ and the dropout rate was chosen on a linear scale from 0 to 1).}
    \LinedItem{The \textit{\textbf{n}} models with can be trained and the random hyperparameter settings, the cost, and the final accuracy after training can be recorded.}
    \LinedItem{Using this generated collection of points, a network designer can then train a separate neural network (or some other Machine Learning model as they see fit) to fit either the cost or the accuracy, in essence to predict the cost/accuracy of the model given the value of each hyperparameter. }
    \LinedItem{The network designer can use this previous model to sample thousands of hyperparameter configurations and generate a surface plot. }
    \LinedItem{Using this surface plot, the designer can ``zoom'' into areas which are predicted by the model to have relatively lower costs. Note that using this surface plot is theoretically equivalent to testing out thousands of configurations using forward propagation in a matter of seconds. } 
    \LinedItem{The network designer could then repeat this process by training another \textit{\textbf{n}} models initialized with random hyperparameter settings \textit{sampled from this smaller value range}, iterating through the process until a model with the least cost is reached. } 
\end{adjustwidth}
\subsection*{Misconceptions}
\paragraph{Misconception 1:} Some may suppose that even with using the proposed method, the designer still needs to sample many configurations of hyperparameters. However, the method ensures that the designer samples fewer points than would be sampled with traditional hyperparameter tuning, yet still has the same result. 

\paragraph{Misconception 2:} Some may suppose that training the neural network to generate the surface plot will take more time than simply trying new configurations. However, even while \textit{\textbf{n}} grows very large, the neural network can be easily trained because there are only two features and one label. 

\subsection*{Generalizations and Implementation Details}
As one might imagine, this hyperparameter tuning process is not restricted to only the hyperparameters we discussed in this example. Rather, the process can be extrapolated to a larger scale containing different hyperparameters or more hyperparameters, allowing designers to more quickly determine the configurations of their models.
\\\\
We remind developers while using this technique to consider the trade off between time and accuracy that is intrinsic to this process, in that the generation of more points expends more computational resources but ultimately results in a better network, while training less points, though timely, may result in a flawed hyperparameter selection. Symbolically, if \textit{\textbf{n}} is too small, then the surface plot will not generalize to untrained points and the neural network will fail to its job. On the other hand, if \textit{\textbf{n}} is too large, then this defeats the purpose of using the proposed method altogether. One potential solution to calibrate the number of models trained is to incrementally decrease the number of configurations tested. For example, a designer could test 100 configurations, zoom in, test another 10 configurations, zoom in, test another 5 configurations, and so on.
\section*{Need For Further Research}
Further experiments can be conducted to determine if the phenomena we observed exist in larger data sets and diverse data sets not limited to cardiovascular disease. The Deep Learning models we trained with 150 epochs can instead be trained for a larger number of iterations to verify the usability of our results before extrapolation to industrial models. Different variations of hyperparameters, such as varying the number of hidden units and dropout rate between each dense layer of a model, can also be tested. This is crucial because in more sophisticated applications of Deep Learning (e.g. convolutional neural networks), different layers of the network learn either ``specific'' or ``general'' features, making it practically impossible for each layer to have the same architecture \cite{architecture}. Thus, more experimentation needs to be done to gauge the usefulness of the procedures when applied to different types of Deep Learning models.
\\\\
In addition, research can be done to gauge the accuracy and reliability of the surface plot neural networks by training many models with random hyperparameters and comparing the results to those predicted by the surface plot neural networks. The significant trade off between surface plot generalizability and computational expense should also be explored to validate or nullify our proposed method of hyperparameter tuning, and to ultimately find the answer to the optimal number of models to train before generating a surface plot.

\subsection*{Acknowledgements}
We would like to thank Mr. Govind Tatachari for his tremendous help and mentorship over the course of this research. We very much appreciate his valuable inputs and feedback.

\section*{Appendix}
\paragraph{Cross-Entropy Cost Function} When training the Deep Learning models, we assess model performance using the Cross-Entropy Cost Function, defined as the following: 
\label{sec:binary}
\[J(\theta)= \frac{1}{m} * \sum^{m}_{i=1}\left[y^{(i)}\log{(h(x^{(i)}))}+(1-y^{(i)})\log{(1-h(x^{(i)}))}\right],\]
where $\theta$ contains the parameters for the neural network, $m$ is the number of examples in the validation data set, $y^{(i)}$ is the ground truth label for example $i$, $x^{(i)}$ is the input for example $i$, $h(x^{(i)})$ is the prediction for example $i$, and $\log$ is the natural logarithm with base $e\approx 2.71828\ldots$

\paragraph{Binary Accuracy}
Let $m$ be the number of examples in the validation data set. Let $x$ be the number of patients who are correctly classified by the Machine Learning model. The Binary Accuracy is then defined as:  
$$\boxed{\text{Binary Accuracy}=\left(100 \cdot \frac{x}{m}\right)\%}.$$
Binary Accuracy ranges from 0\% to 100\%, where 100\% corresponds to perfect model performance on validation data.

\end{document}